\documentclass[letterpaper, 10 pt, conference]{ieeeconf}  
\IEEEoverridecommandlockouts                              
\usepackage{graphics}    
\usepackage{times}       
\usepackage{amsmath}     
\usepackage{amssymb}     
\usepackage{graphicx}
\usepackage[noline, ruled]{algorithm2e}
\usepackage[noend]{algpseudocode}
\usepackage{subfigure}
\usepackage{subfloat}
\usepackage{booktabs}
\usepackage{xspace}
\usepackage{pifont}
\usepackage{color}
\usepackage{hyperref}

\usepackage[font=small]{caption}

\def\secref#1{Sec.~\ref{#1}}
\def\figref#1{Fig.~\ref{#1}}
\def\tabref#1{Tab.~\ref{#1}}
\def\eqref#1{Eq.~(\ref{#1})}
\def\algref#1{Alg.~\ref{#1}}

\renewcommand{\vec}[1]{\mathbf{#1}}	
\newcommand{\mat}[1]{\mathbf{#1}}  	
\newcommand{\set}[1]{\mathcal{#1}} 	

\newcommand{\RR}{\mathbb{R}} 

\newcommand\etal{\emph{et al.}}

\newcommand{\rgbd}{\mbox{RGB-D}\xspace}

\newif\iffinalversion
\finalversionfalse

\iffinalversion

\else

\fi

\usepackage{array}
\newcolumntype{L}[1]{>{\raggedright\let\newline\\\arraybackslash\hspace{0pt}}m{#1}}
\newcolumntype{C}[1]{>{\centering\let\newline\\\arraybackslash\hspace{0pt}}m{#1}}
\newcolumntype{R}[1]{>{\raggedleft\let\newline\\\arraybackslash\hspace{0pt}}m{#1}}

\newcommand\Tstrut{\rule{0pt}{2.6ex}}         

\makeatletter
\DeclareRobustCommand\onedot{\futurelet\@let@token\@onedot}
\def\@onedot{\ifx\@let@token.\else.\null\fi\xspace}

\def\ie{i.e\onedot}

\def\etal{\emph{et al}\onedot}
\makeatother

\title{\LARGE \bf SuMa++: Efficient LiDAR-based Semantic SLAM}

\author{Xieyuanli Chen \and Andres Milioto\and  Emanuele Palazzolo \and  Philippe Gigu\`{e}re \and
		Jens Behley \and Cyrill Stachniss
  \thanks{X.Chen, A. Milioto, E. Palazzolo, J. Behley, and C. Stachniss are with the University of Bonn, Germany. Philippe Gigu\`{e}re is with the Laval University, Qu{\'e}bec, Canada. This work has partly been supported by the German Research Foundation under Germany's Excellence Strategy, EXC-2070 - 390732324 (PhenoRob) and under grant number BE 5996/1-1 as well as by the Chinese Scholarship Committee.
  }%
}

\begin{document}
\maketitle
\thispagestyle{empty}
\pagestyle{empty}

\begin{abstract}

Reliable and accurate localization and mapping are key components of most autonomous systems.
Besides geometric information about the mapped environment, the semantics plays an important role to enable intelligent navigation behaviors.
In most realistic environments, this task is particularly complicated due to dynamics caused by moving objects, which can corrupt the mapping step or derail localization.
In this paper, we propose an extension of a recently published surfel-based mapping approach exploiting three-dimensional laser range scans by integrating semantic information to facilitate the mapping process.
The semantic information is efficiently extracted by a fully convolutional neural network and rendered on a spherical projection of the laser range data.
This computed semantic segmentation results in point-wise labels for the whole scan, allowing us to build a semantically-enriched map with labeled surfels.
This semantic map enables us to reliably filter moving objects, but also improve the projective scan matching via semantic constraints.
Our experimental evaluation on challenging highways sequences from KITTI dataset with very few static structures and a large amount of moving cars shows the advantage of our semantic SLAM approach in comparison to a purely geometric, state-of-the-art approach.

\end{abstract}

\section{Introduction}
\label{sec:intro}

\begin{figure}[t]
 \centering
 \includegraphics[width=0.95\linewidth]{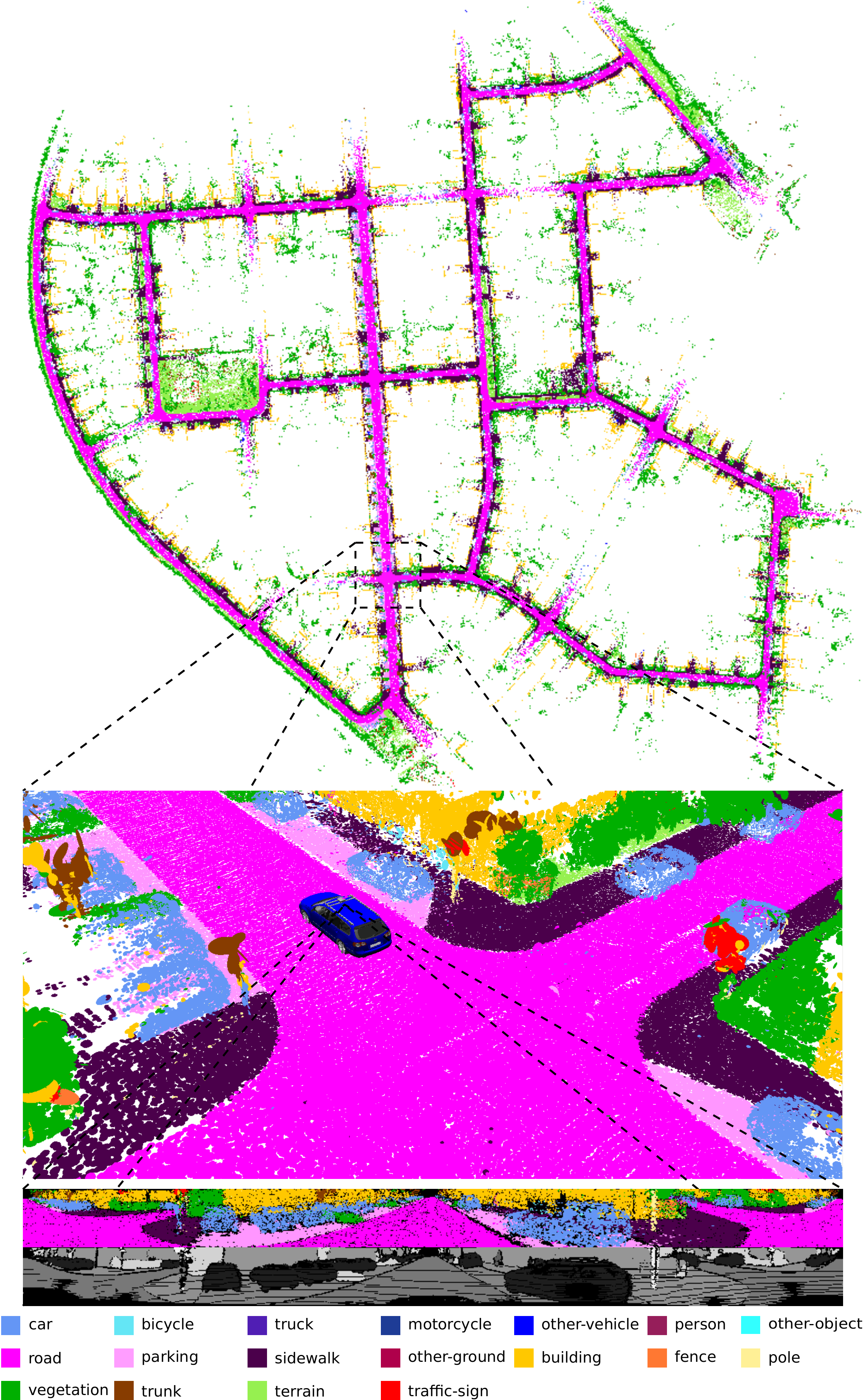}
 \caption{Semantic map of the KITTI dataset generated with our approach using only LiDAR scans.
 The map is represented by surfels that have a class label indicated by the respective color. 
 Overall, our semantic SLAM pipeline is able to provide high-quality semantic maps with higher metric accuracy than its non-semantic counterpart.}
\label{fig:motivation}
\vspace{-0.4cm}
\end{figure}

Accurate localization and reliable mapping of unknown environments are fundamental for most autonomous vehicles.
Such systems often operate in highly dynamic environments, which makes the generation of consistent maps more difficult.
Furthermore, semantic information about the mapped area is needed to enable intelligent navigation behavior.
For example, a self-driving car must be able to reliably find a location to legally park, or pull over in places where a safe exit of the passengers is possible -- even in locations that were never seen, and thus not accurately mapped before.

In this work, we propose a novel approach to simultaneous localization and mapping (SLAM) able to generate such semantic maps by using three-dimensional laser range scans.
Our approach exploits ideas from a modern LiDAR SLAM pipeline \cite{behley2018rss} and incorporates semantic information obtained from a semantic segmentation generated by a Fully Convolutional Neural Network (FCN)~\cite{milioto2019iros}. 
This allows us to generate high-quality semantic maps, while at the same time improve the geometry of the map and the quality of the odometry.

The FCN provides class labels for each point of the laser range scan. We perform highly efficient processing of the point cloud by first using a spherical projection.
Classification results on this two-dimensional spherical projection are then back-projected to the three-dimensional point cloud.
However, the back-projection introduces artifacts, which we reduce by a two-step process of an erosion followed by depth-based flood-fill of the semantic labels.
The semantic labels are then integrated into the surfel-based map representation and exploited to better register new observations to the already built map.
We furthermore use the semantics to filter moving objects by checking semantic consistency between the new observation and the world model when  updating the map. 
In this way, we reduce the risk of integrating dynamic objects into the map.
\figref{fig:motivation} shows an example of our semantic map representation. The semantic classes are generated by the FCN of Milioto \etal~\cite{milioto2019iros}, which was trained using the SemanticKITTI dataset by Behley \etal~\cite{behley2019iccv}.

The main contribution of this paper is an approach to integrate semantics into a surfel-based map representation and a method to filter dynamic objects exploiting these semantic labels.
In sum, we claim that we are (i) able to accurately map an environment especially in situations with a large number of moving objects and we are (ii) able to achieve a better performance than the same mapping system simply removing possibly moving objects in general environments, including urban, countryside, and highway scenes.
We experimentally evaluated our approach on challenging sequences of  KITTI~\cite{geiger2012cvpr} and show superior performance of our semantic surfel-mapping approach, called \emph{SuMa++}, 
compared to purely geometric surfel-based mapping and compared mapping removing all potentially moving objects based on class labels. The source code of our approach is  available at:\\
\url{https://github.com/PRBonn/semantic_suma/}

\section{Related Work}
\label{sec:related}

Odometry estimation and SLAM are classical topics in robotics with a large body of scientific work covered by several overview articles~\cite{bresson2017itiv, cadena2016tro, stachniss2016handbook-slamchapter}. 
Here, we mainly concentrate on related work for semantic SLAM based on learning approaches and dynamic scenes.
	
Motivated by the advances of deep learning and Convolutional Neural Networks~(CNNs) for scene understanding, there have been many semantic SLAM techniques exploiting this information using cameras~\cite{brasch2018iros, tateno2017cvpr}, cameras + IMU data~\cite{bowman2017icra}, stereo cameras~\cite{ganti2018arxiv, li2018eccv, lianos2018eccv, vineet2015icra, yang2017iros}, or \rgbd sensors~\cite{bescos2018ral, mccormac2018threedv, mccormac2017icra, ruenz2018ismar, salas2013cvpr, suenderhauf2017iros, yu2018iros}. Most of these approaches were only applied indoors and use either an object detector or a semantic segmentation of the camera image.
In contrast, we only use laser range data and exploit information from a semantic segmentation operating on depth images generated from LiDAR scans.

There exists also a large body of literature tackling localization and mapping changing environments, for example by filtering moving objects~\cite{kuemmerle2013icra}, considering residuals in matching~\cite{palazzolo2019iros}, or by exploiting sequence information~\cite{vysotska2016ral}.
To achieve outdoor large-scale semantic SLAM, one can also combine 3D LiDAR sensors with RGB cameras. Yan \etal~\cite{yan2014threedv} associate 2D images and 3D points to improve the segmentation for detecting moving objects. 
Wang and Kim~\cite{wang2017iros} use images and 3D point clouds from the KITTI dataset~\cite{geiger2012cvpr} to jointly estimate road layout and segment urban scenes semantically by applying a relative location prior. 
Jeong \etal~\cite{jeong2018esa, jeong2018sensors} also propose a multi-modal sensor-based semantic 3D mapping system to improve the segmentation results in terms of the intersection-over-union (IoU) metric, in large-scale environments as well as in environments with few features. Liang \etal~\cite{liang2018eccv} propose a novel 3D object detector that can exploit both LiDAR and camera  data to perform accurate object localization. All of these approaches focus on combining 3D LiDAR and cameras to improve the object detection, semantic segmentation, or 3D reconstruction.

The recent work by Parkison \etal~\cite{parkison2018bmvc} develops a point cloud registration algorithm by directly incorporating image-based semantic information into the estimation of the relative transformation between two point clouds. 
A subsequent work by Zaganidis \etal~\cite{zaganidis2018ral} realizes both LiDAR combined with images and LiDAR only semantic 3D point cloud registration. Both approaches use semantic information to improve pose estimation, but they cannot be used for online operation because of the long processing time. 

The most similar approaches to the one proposed in this paper are Sun \etal~\cite{sun2018ral} and Dub\'e \etal~\cite{dube2018rss}, which realize semantic SLAM using only a single LiDAR sensor. Sun \etal~\cite{sun2018ral} present a semantic mapping approach, which is formulated as a sequence-to-sequence encoding-decoding problem. 
Dub\'{e} \etal~\cite{dube2018rss} propose an approach called SegMap, which is based on segments extracted from the point cloud and assigns semantic labels to them. They mainly aim at extracting meaningful features for the global retrieval and multi-robot collaborative SLAM with very limited types of semantic classes.
In contrast to them, we focus on generating a semantic map with an abundance of semantic classes and using these semantics to filter outliers caused by dynamic objects, like moving vehicles and humans, to improve both mapping and odometry accuracy.

\section{Our Approach}
\label{sec:main}

\begin{figure*}[!t]
\vspace{0.2cm}
  \centering
 \includegraphics[width=0.95\linewidth]{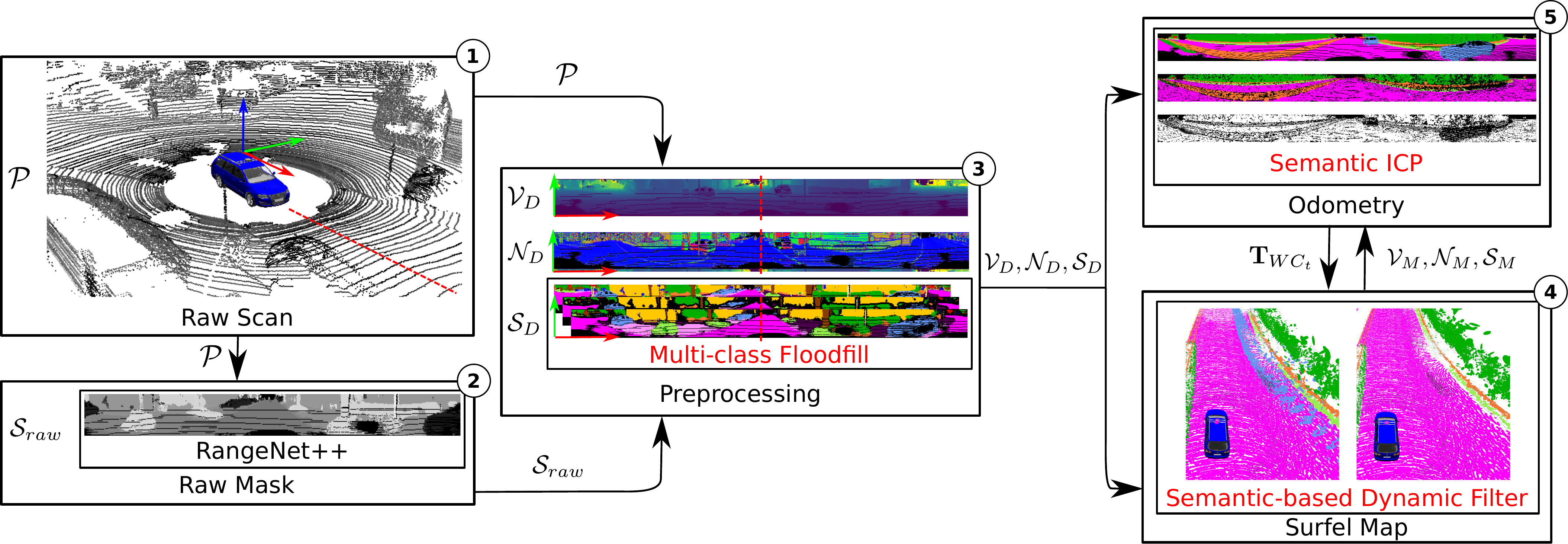}
 \caption{Pipeline overview of our proposed approach. We integrate semantic predictions into the SuMa pipeline in a compact way: (1) The input is only the LiDAR scan $\set{P}$. (2) Before processing the raw point clouds $\set{P}$, we first use a semantic segmentation from RangeNet++ to predict the semantic label for each point and generate a raw semantic mask $\set{S}_{\mathrm{raw}}$. (3) Given the raw mask, we generate an refined semantic map $\set{S}_D$ in the preprocessing module using multi-class flood-fill. (4) During the map updating process, we add a dynamic detection and removal module which checks the semantic consistency between the new observation $\set{S}_D$ and the world model $\set{S}_M$ and remove the outliers. (5) Meanwhile, we add extra semantic constraints into the ICP process to make it more robust to outliers.}
\label{fig:pipeline}
\vspace{-0.4cm}
\end{figure*}

The foundation of our semantic SLAM approach is our Surfel-based Mapping~(SuMa)~\cite{behley2018rss} pipeline, which we extend by integrating semantic information provided by a semantic segmentation using the FCN RangeNet++~\cite{milioto2019iros} as illustrated in \figref{fig:pipeline}.
The point-wise labels are provided by RangeNet++ using spherical projections of the point clouds. This information is then used to filter dynamic objects and to add semantic constraints to the scan registration, which improves the robustness and accuracy of the pose estimation by SuMa.

\subsection{Notation} 

We denote the transformation of a point $\vec{p}_A$ in coordinate frame $A$ to a point $\vec{p}_B$ in coordinate frame $B$ by $\mat{T}_{BA} \in \RR^{4\times4}$, such that $\vec{p}_B = \mat{T}_{BA} \vec{p}_A$.
Let $\mat{R}_{BA} \in \mathrm{SO}(3)$ and $\vec{t}_{BA} \in \RR^{3}$ denote the corresponding rotational and translational part of transformation $\mat{T}_{BA}$.

We call the coordinate frame at timestep $t$ as $C_t$.
Each variable in coordinate frame $C_t$ is associated to the world frame $W$ by a pose $\mat{T}_{WC_t} \in \RR^{4 \times 4}$, transforming the observed point cloud into the world coordinate frame.

\subsection{Surfel-based Mapping}

Our approach relies on SuMa, but we only summarize here the main steps relevant to our approach and refer for more details to the original paper \cite{behley2018rss}.
SuMa first generates a spherical projection of the point cloud $\set{P}$ at timestep $t$, the so-called vertex map  $\set{V}_D$, which is then used to generate a corresponding normal map $\set{N}_D$. Given this information, SuMa determines via projective ICP in a rendered map view $\set{V}_M$ and $\set{N}_M$ at timestep $t-1$ the pose update $\mat{T}_{C_{t-1}C_{t}}$ and consequently $\mathbf{T}_{WC_t}$ by chaining all pose increments. 

The map is represented by surfels, where each surfel is defined by a position $\vec{v}_s \in \RR^3$, a normal $\vec{n}_s \in \RR^3$, and a radius $r_s \in \RR$.
Each surfel additionally carries two timestamps: the creation timestamp $t_c$ and the timestamp $t_u$ of its last update by a measurement. Furthermore, a stability log odds ratio $l_s$ is maintained using a binary Bayes Filter \cite{thrun2005probrobbook} to determine if a surfel is considered stable or unstable.
SuMa also performs loop closure detection with subsequent pose-graph optimization to obtain globally consistent maps.

\subsection{Semantic Segmentation}

For each frame, we use RangeNet++~\cite{milioto2019iros} to predict a semantic label for each point and generate a semantic map $\set{S}_D$.
RangeNet++ semantically segments a range image generated by a spherical projection of each laser scan. 
Briefly, the network is based on the SqueezeSeg architecture proposed by Wu~\etal \cite{wu2018icra-scnn} and uses a DarkNet53 backbone proposed by Redmon \etal~\cite{redmon2018arxiv} to improve results by using more parameters, while keeping the approach real-time capable.  For more details about  the semantic segmentation approach, we refer to the paper of Milioto \etal~\cite{milioto2019iros}.
The availability of point-wise labels in the field of view of the sensor makes it also possible to integrate the semantic information into the map.
To this end, we add for each surfel the inferred semantic label~$y$ and the corresponding probability of that label from the semantic segmentation.

\subsection{Refined Semantic Map}

\begin{algorithm}[t]

\DontPrintSemicolon
\SetKw{Break}{break}
\KwIn{semantic mask $\set{S}_{\mathrm{raw}}$ and the corresponding vertex map $\set{V}_D$}
\KwResult{refined mask $\set{S}_D$}
  Let $\set{N}_{\vec{s}}$ be the set of neighbors of pixel $\vec{s} \in \set{S}$ within a filter kernel of size $d$.\;
  $\theta$ is the rejection threshold.\;
  $\vec{0}$ represents an empty pixel with label 0.\;
  \ForEach{$\vec{s}_\vec{u} \in \set{S}_{\mathrm{raw}}$}
  {  	          
    Let $\set{S}_{\mathrm{raw}}^{\mathrm{eroded}}(\vec{u}) = \vec{s}_\vec{u}$ \; 
    \ForEach{$\vec{n} \in \set{N}_{\vec{s}_\vec{u}}$}
    {
      \If{ $y_{\vec{s}_\vec{u}} \neq y_{\vec{n}}$ } 
    {
      Let $\set{S}_{\mathrm{raw}}^{\mathrm{eroded}}(\vec{u}) = \vec{0}$ \;
      \Break \;
    }   
    }
  }

  \ForEach{$\vec{s}_\vec{u} \in \set{S}_{\mathrm{raw}}^{\mathrm{eroded}}$}
  {
    Let $\set{S}_D(\vec{u}) = \vec{s}_\vec{u}$ \;
    \ForEach{$\vec{n} \in \set{N}_{\vec{s}_\vec{u}}$}
    {
      \If{ $|| \vec{u} - \vec{u}_{\vec{n}} || < \theta \cdot ||\vec{u}||$} 
    {
      Let $\set{S}_D(\vec{u}) = \vec{n}$  \textbf{if} $y_{\vec{s}_\vec{u}} = 0$\;
      \Break \;
    }
    }
  }

\caption{\textbf{flood-fill} for refining $\set{S}_D$.}

\label{alg:flood-fill}
\end{algorithm}

Due to the projective input and the blob-like outputs produced as a by-product of in-network down-sampling of RangeNet++, we have to deal with errors of the semantic labels, when the labels are re-projected to the map.
To reduce these errors, we use a flood-fill algorithm, summarized in~\algref{alg:flood-fill}. It is inside the preprocessing module, which uses depth information from the vertex map $\set{V}_D$ to refine the semantic mask $\set{S}_D$.

The input to the flood-fill is the raw semantic mask $\set{S}_{\mathrm{raw}}$ generated by the RangeNet++ and the corresponding vertex map $\set{V}_D$. The value of each pixel in the mask $\set{S}_{\mathrm{raw}}$ is a semantic label. The corresponding pixel in the vertex map contains the 3D coordinates of the nearest 3D point in the LiDAR coordinate system. The output of the approach is the refined semantic mask $\set{S}_D$.
 
Considering that the prediction uncertainty of object boundaries is higher than that of the center of an object~\cite{li2017cvpr}, we use the following two steps in the flood-fill.
The first step is an erosion that removes pixels where the neighbors within a kernel of size $d$ show at least one different semantic label resulting in the eroded mask $\set{S}_{\mathrm{raw}}^{\mathrm{eroded}}$. 
Combining this mask with the depth information generated from the vertex map $\set{V}_D$, we then fill-in the eroded mask. 
To this end, we set the label of an empty boundary pixel to the neighboring labeled pixels if the distances of the corresponding points are consistent, \ie, less than a threshold $\theta$.

\begin{figure}[t]
\vspace{0.2cm}
  \centering
 \includegraphics[width=0.99\linewidth]{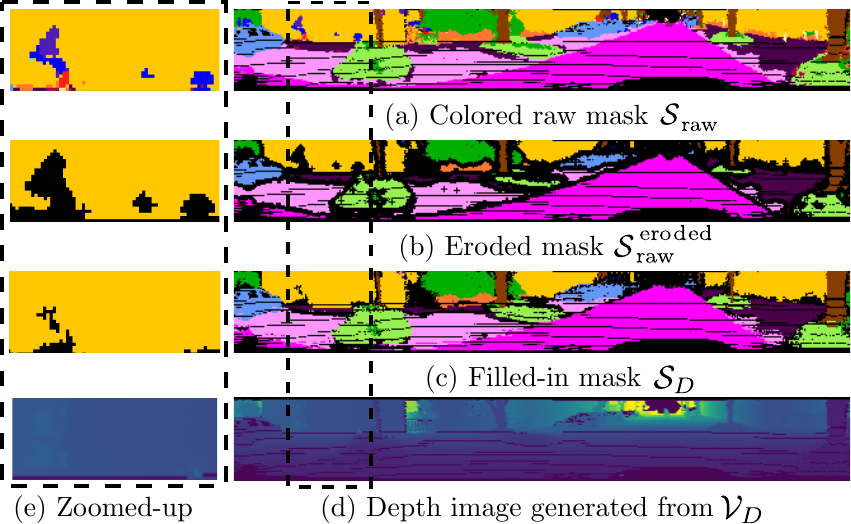}
 \caption{Visualization of the processing steps of the proposed flood-fill 
algorithm. Given the (a) raw semantic map $\mathcal{S}_{\text{raw}}$, we 
first use an erosion to remove boundary labels and small areas of wrong 
labels resulting in (b) the eroded mask $\mathcal{S}^{eroded}_{\text{raw}}$. 
(c) We then finally fill-in eroded labels with neighboring labels to get 
a more consistent result $\mathcal{S}_{\mathcal{D}}$. 
Black points represent empty pixels with label 0.
(d) Shows the depth and (e) the details inside the areas with the dashed borders.}
\label{fig:flood-fill}
\vspace{-0.4cm}
\end{figure}

\figref{fig:flood-fill} shows the intermediate steps of this algorithm.
Note that the filtered semantic map does contain less artifacts compared to the raw predictions. For instance, the wrong labels on the wall of the building are mostly corrected, which is illustrated in \figref{fig:flood-fill}(e).

\subsection{Filtering Dynamics using Semantics}
\label{sec:dynamic-filter}

Most existing SLAM systems rely on geometric information to represent the environment and associate observations to the map. 
They work well under the assumption that the environment is mostly static. 
However, the world is usually dynamic, especially when  considering driving scenarios, and several of the traditional approaches fail to account for dynamic scene changes caused by moving objects.
Therefore, moving objects can cause wrong associations between observations and the map in such situations, which must be treated carefully.
Commonly, SLAM approaches use some kind of outlier rejection, either by directly filtering the observation or by building map representations that filter out changes caused by moving objects.

\begin{figure}[t]
\vspace{0.1cm}
  	\centering
 	\subfigure[SuMa]{\includegraphics[width=0.325\linewidth]{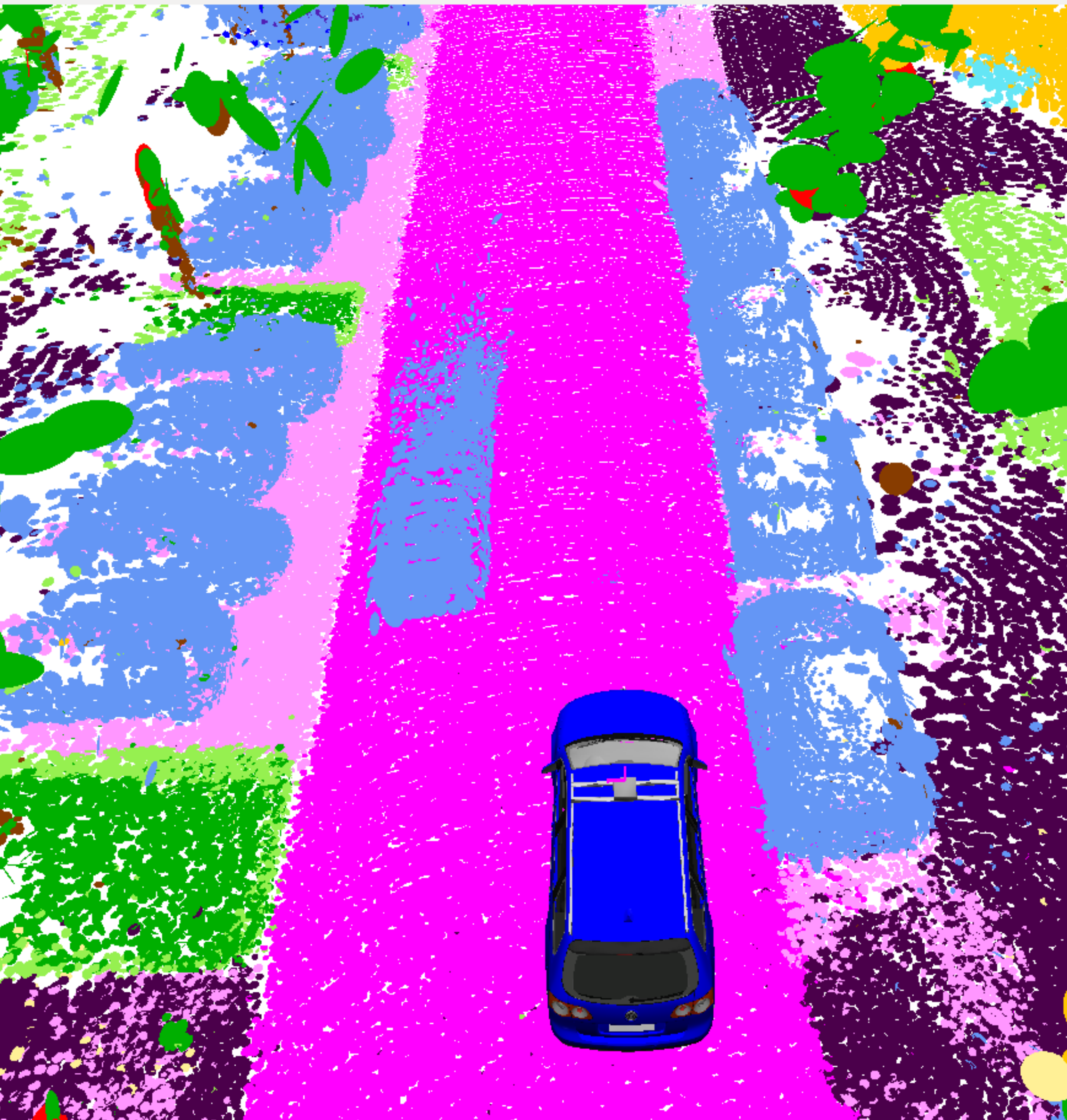}}
 	\subfigure[SuMa++]{\includegraphics[width=0.325\linewidth]{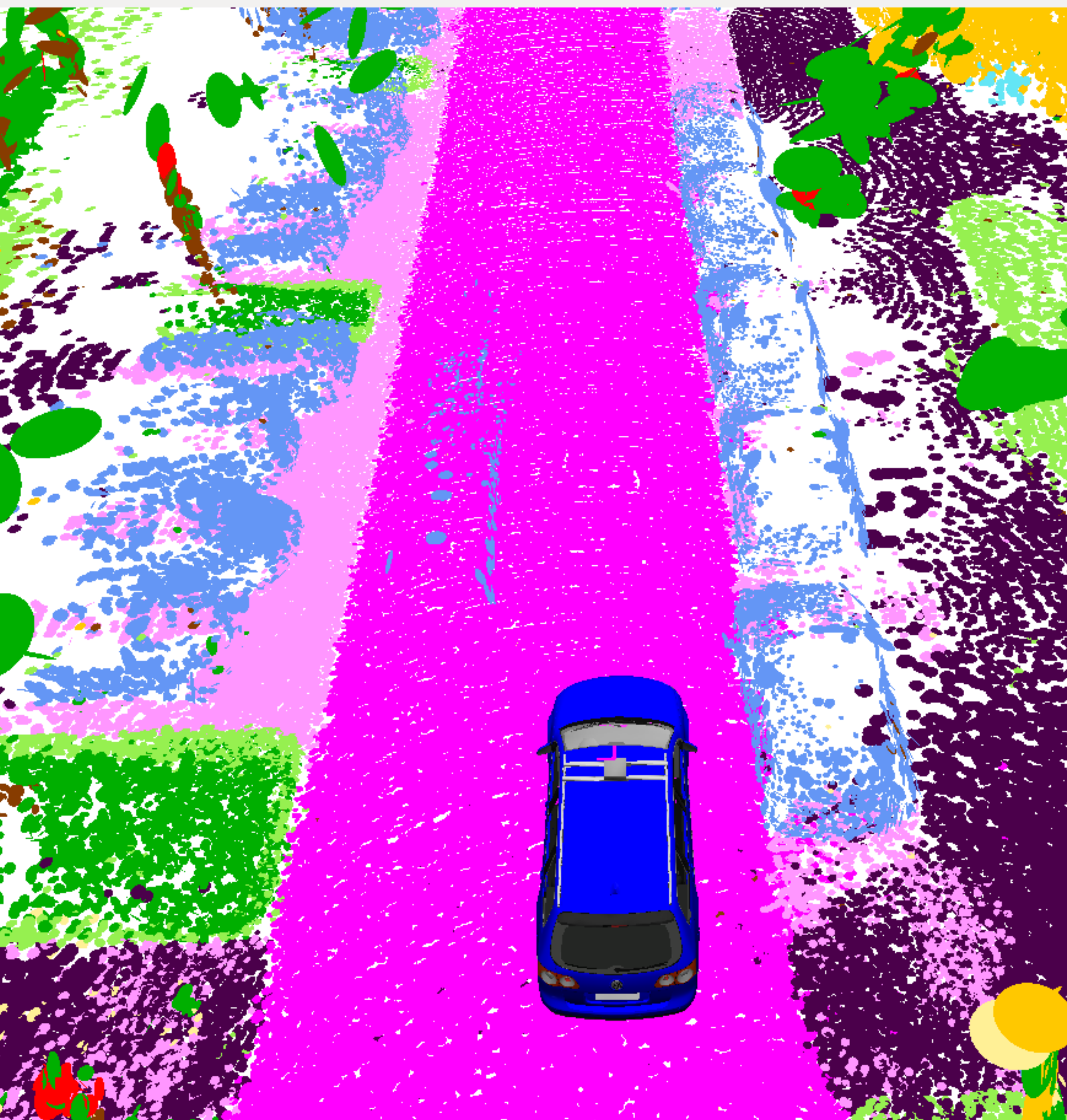}}
 	\subfigure[SuMa\_nomovable]{\includegraphics[width=0.325\linewidth]{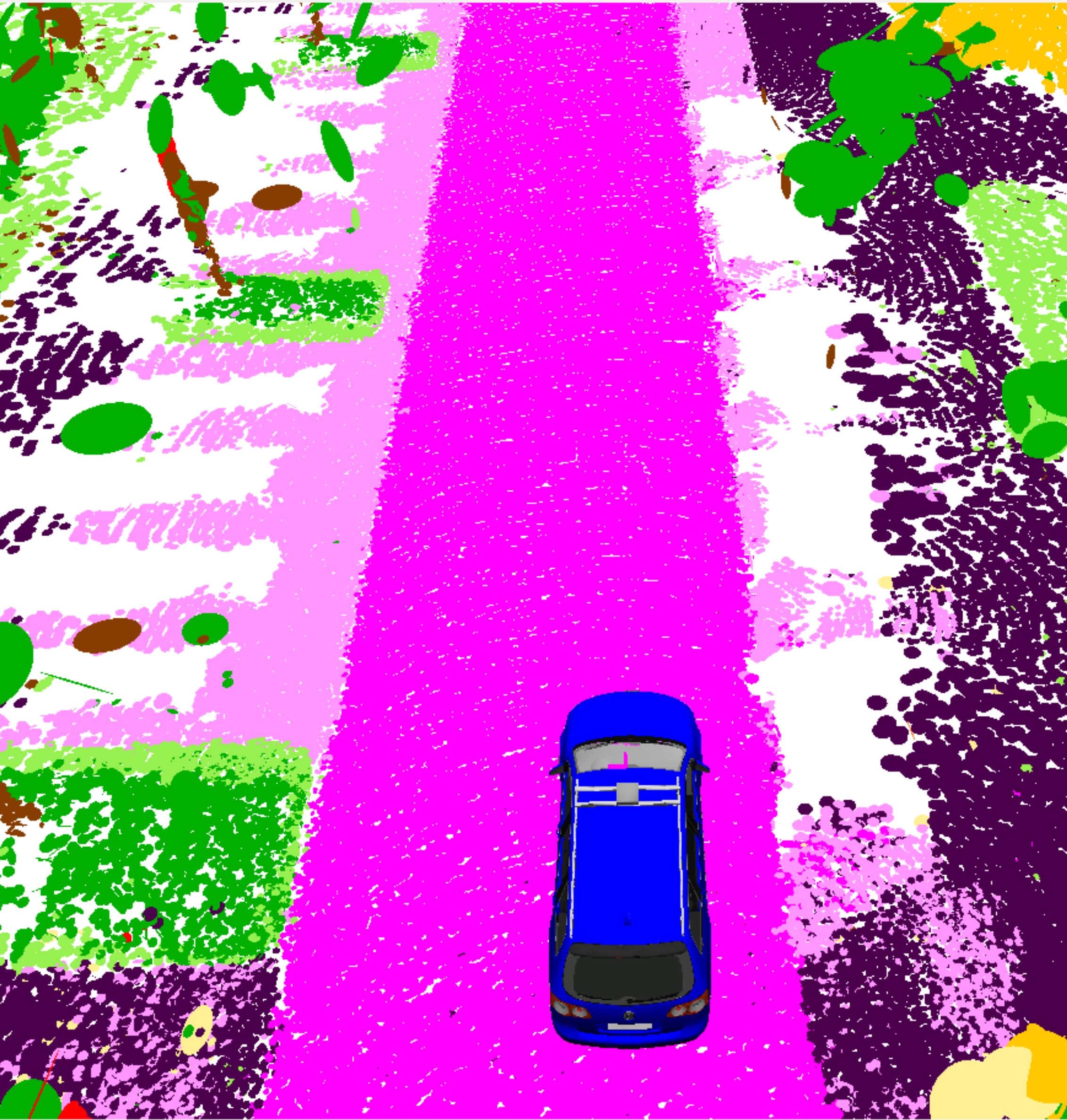}}
 	\caption{Effect of the proposed filtering of dynamics. For all figures, we show the color of the corresponding label, but note that SuMa does not use the semantic information. (a) Surfels generated by SuMa; (b) our method; (c) remove all potentially moving objects.}
\label{fig:dynamic_detection}
\vspace{-0.4cm}
\end{figure}

In our approach, we exploit the labels provided by the semantic segmentation to handle moving objects.
More specifically,  we filter dynamics by checking semantic consistency between the new observation $\set{S}_D$ and the world model $\set{S}_M$, when we update the map. 
If the labels are inconsistent, we assume those surfels belong to an object that moved between the scans. 
Therefore, we add a penalty term to the computation of the stability term in the recursive Bayes filter. 
After several observations, we can remove the unstable surfels. In this way, we achieve a  detection of dynamics and finally a removal.

More precisely, we penalize that surfel by giving a penalty $\mathrm{odds}(p_{\mathrm{penalty}})$ to its stability log odds ratio $l_s$, which will be updated as follows:
\begin{align}
l^{(t)}_s &= l^{(t-1)}_s \nonumber \\
& \quad + \mathrm{odds}\left(p_{\mathrm{stable}} \exp\left(-\frac{\alpha^2}{\sigma_{\alpha}^2}\right)\exp\left(-\frac{d^2}{\sigma_{d}^2}\right)\right) \nonumber \\
& \quad - \mathrm{odds}(p_{\mathrm{prior}}) - \mathrm{odds}(p_{\mathrm{penalty}}),
\label{eq:stability}
\end{align}
where $\mathrm{odds}(p) = \log(p (1-p)^{-1})$ and $p_{\mathrm{stable}}$ and $p_{\mathrm{prior}}$ are probabilities for a stable surfel given a compatible measurement and the prior probability, respectively. The terms $\exp(-x^2\sigma^{-2})$ are used to account for noisy measurements, where $\alpha$ is the angle between the surfel's normal $\vec{n}_s$ and the normal of the measurement to be integrated, and $d$ is the distance of the measurement in respect to the associated surfel. 
The measurement normal is taken from $\set{N}_D$ and the correspondences from the frame-to-model ICP, see \cite{behley2018rss} for more details.

Instead of using semantic information, Pomerleau \etal~\cite{pomerleau2014icra} proposes a method to infer the dominant motion patterns within the map by storing the time history of velocities. 
In contrast to our approach, their method requires a given global map to estimate the velocities of points in the current scan. Furthermore, their robot pose estimate is assumed to be rather accurate.

In \figref{fig:dynamic_detection}, we illustrate the effect of our filtering method compared to naively removing all surfels from classes corresponding to movable objects. 
When utilizing the naive method, surfels on parked cars are removed, even though these might be valuable features for the incremental pose estimation.
With the proposed filtering, we can effectively remove the dynamic outliers and obtain a cleaner semantic world model, while keeping surfels from static objects, e.g., parked cars.
These static objects are valuable information for the ICP scan registration and simply removing them can lead to failures in scan registration due to missing correspondences.

\subsection{Semantic ICP}

To further improve the pose estimation using the frame-to-model ICP, we also add semantic constraints to the optimization problem, which helps to reduce the influence of outliers.
Our error function to minimize for ICP is given by:
\begin{align}
 E(\set{V}_D, \set{V}_M, \set{N}_M) &= \sum_{\vec{u} \in \set{V}_D} w_\vec{u}  \underbrace{ \vec{n}_\vec{u}^\top \left(\mat{T}^{(k)}_{C_{t-1}C_{t}}\vec{u} - \vec{v}_\vec{u}\right)^2}_{r_{\vec{u}}} , \label{eq:objective}
\end{align}
where each vertex $\vec{u} \in \set{V}_D$ is projectively associated to a reference vertex $\vec{v}_\vec{u} \in \set{V}_M$ and its normal $\vec{n}_\vec{u} \in \set{N}_M$ via
\begin{align}
\vec{v}_\vec{u} &= \set{V}_M\left(\Pi\left(\mat{T}^{(k)}_{C_{t-1}C_{t}}\vec{u}\right)\right), \\ 
\vec{n}_\vec{u} &= \set{N}_M\left(\Pi\left(\mat{T}^{(k)}_{C_{t-1}C_{t}}\vec{u}\right)\right),
\end{align}
$r_{\vec{u}}$ and $w_\vec{u}$ are the corresponding residual and weight, respectively.

For the minimization, we use Gauss-Newton and determine increments~$\delta$ by iteratively solving:
\begin{align}
 \delta &= \left(\mat{J}^\top \mat{W} \mat{J}\right)^{-1}\mat{J}^\top\mat{W}\vec{r},
\end{align}
where $\mat{W} \in \RR^{n\times n}$ is a diagonal matrix containing weights $w_\vec{u}$ for each residual $r_\vec{u}$, $\vec{r} \in \RR^n$ is the stacked residual vector, and $\mat{J} \in \RR^{n\times 6}$ the Jacobian of $\vec{r}$ with respect to the increment $\delta$.  
Besides the hard association and weighting by a Huber norm, we add extra constraints from higher level semantic scene understanding to weight the residuals. 
In this way, we can combine semantics with geometric information to make the ICP process more robust to outliers.

\begin{figure}[t]
\vspace{0.2cm}
 \centering
 \subfigure[Semantic map of current observation $\set{S}_D$]{\includegraphics[width=0.99\linewidth]{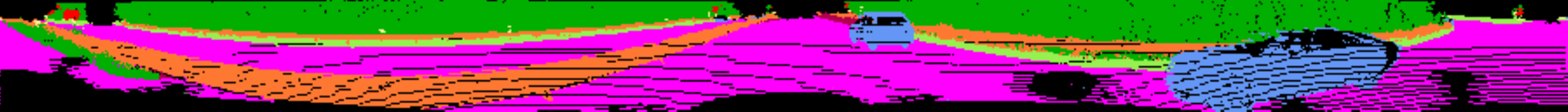}\label{fig:semicp_data}}
 \subfigure[Semantic map of world model $\set{S}_M$]{\includegraphics[width=0.99\linewidth]{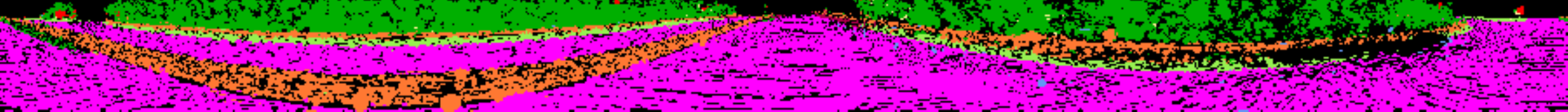}\label{fig:semicp_model}}
 \subfigure[Visualized weights map]{\includegraphics[width=0.99\linewidth]{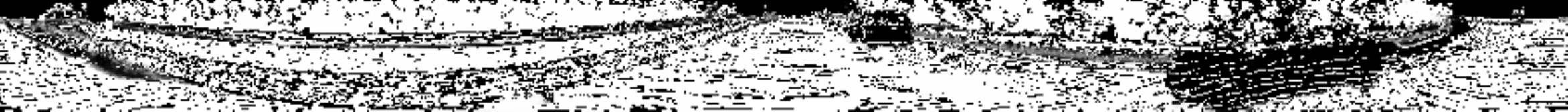}\label{fig:semicp_weight}}
 \caption{Visualization of Semantic ICP: (a) semantic map $\set{S}_D$ for the current laser scan, (b) corresponding semantic map $\set{S}_M$ rendered from the model, 
 (c) the weight map during the ICP. The darker the pixel, the lower is the weight of the corresponding pixel.}
\label{fig:semantic_ICP}
\vspace{-0.4cm}
\end{figure}

Within ICP, we compute the weight $w_{\vec{u}}^{(k)}$ for the residual $r_{\vec{u}}^{(k)}$ in iteration $k$ as follows:
\begin{align}
w_{\vec{u}}^{(k)} &= \rho_{\text{Huber}}\left(r_{\vec{u}}^{(k)}\right)C_{\text{semantic}}(\set{S}_D(\vec{u}), \set{S}_M(\vec{u})) \nonumber\\ 
&\quad\quad \mathbb{I}\Big\{l^{(k)}_s \ge l_{stable}\Big\}, 
\end{align}
where $\rho_{\text{Huber}}(r)$ corresponds to the Huber norm, given by:
\begin{align}
 \rho_{\text{Huber}}(r) &= \left\{\begin{array}{cl} 1 & \text{, if } |r| < \delta \\ \delta |r|^{-1} & \text{, otherwise.} \end{array}\right. .
\end{align}
For semantic compatibility $C_{\text{semantic}}\big((y_\vec{u},P_\vec{u}), (y_{v_\vec{u}},P_{v_\vec{u}})\big)$, the term is defined as:
\begin{align}
C_{\text{semantic}}(\cdot,\cdot) &= \left\{\begin{array}{cl} P(y_\vec{u}|\vec{u}) & \text{, if } y_\vec{u} = y_{v_\vec{u}} \\  1-P(y_\vec{u}|\vec{u}) & \text{, otherwise.} \end{array}\right. ,
\end{align}
which is using the certainty of the predicted label to weight the residual.
By $\mathbb{I}\{a\}$, we denote the indicator function that is $1$ if the argument $a$ is true, and $0$ otherwise.

\figref{fig:semantic_ICP} shows the weighting for a highway scene with two moving cars visible in the scan, see \figref{fig:semicp_data}.
Note that our filtering of dynamics using the semantics, as described in \secref{sec:dynamic-filter}, removed the moving cars from the map, see \figref{fig:semicp_model}.
Therefore, we can also see a low weight corresponding to lower intensity in \figref{fig:semicp_weight}, since the classes of the observation and the map disagree.

\section{Experimental Evaluation}
\label{sec:exp}

\begin{figure}
 \centering
 \vspace{0.1cm} 
 \subfigure[SuMa]{\includegraphics[height=4.6cm]{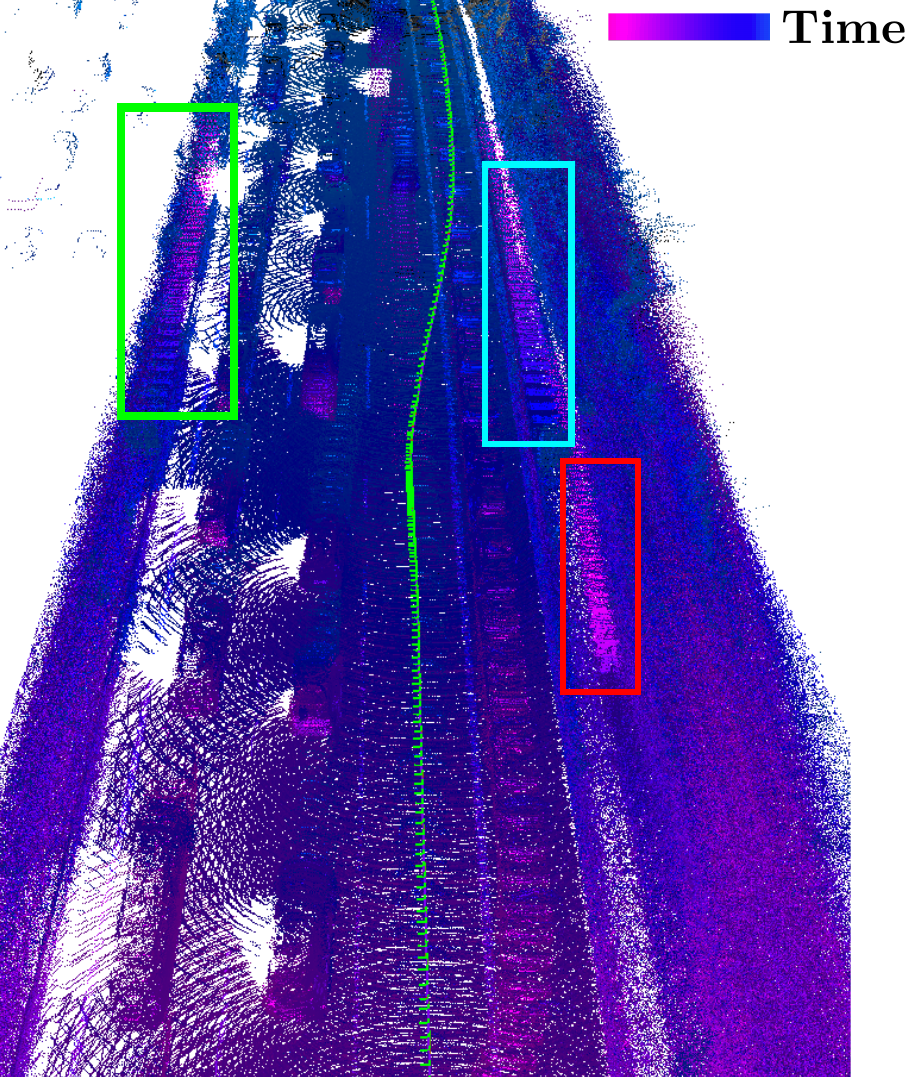}}
 \hspace{0.3cm}
 \subfigure[SuMa++]{\includegraphics[height=4.6cm]{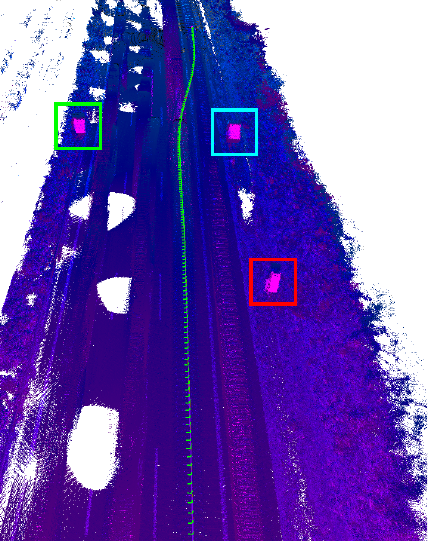}}
\subfigure[Corresponding front-view camera image]{\includegraphics[width=0.93\linewidth]{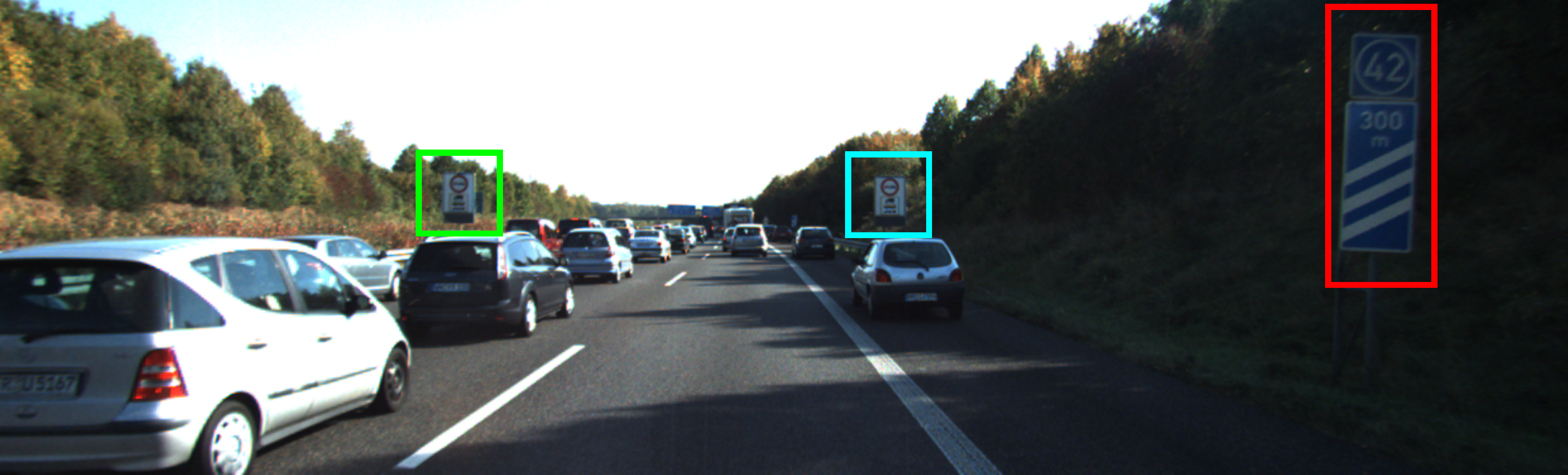}}
\subfigure[Relative translation error plot for each time step]{\includegraphics[width=0.93\linewidth]{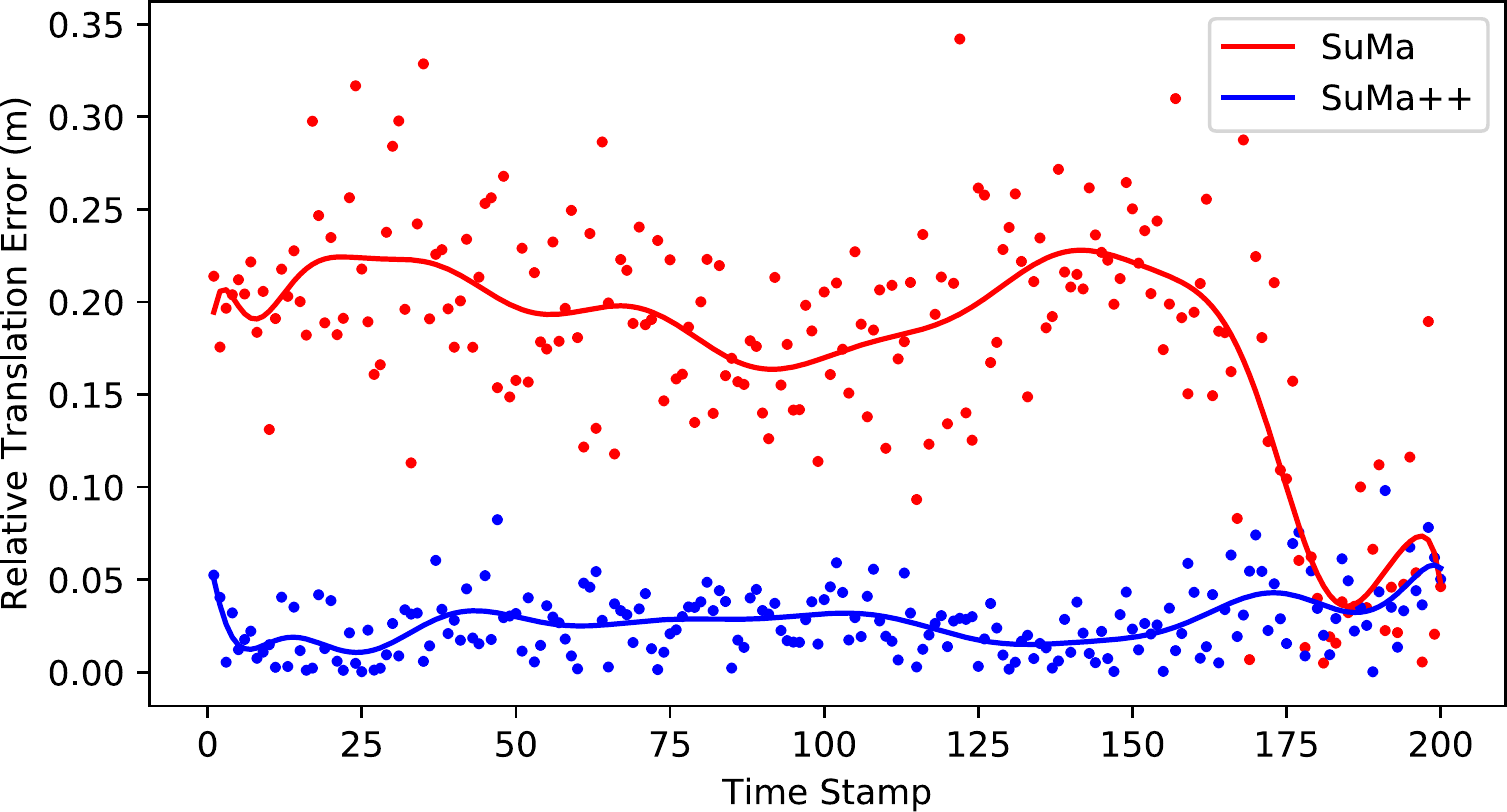}}
 \caption{Qualitative results. (a) SuMa without semantics fails to correctly estimate the motion of the sensor due to the consistent movement of cars in the vicinity of the sensor. The frame-to-model ICP locks to consistently moving cars leading to the map inconsistencies, highlighted by rectangles. (b) By incorporating semantics, we are able to correctly estimate the sensor's movement and therefore get a more consistent map of the environment and a better estimate of sensor pose via ICP. The color of the 3D points refers to the timestamp when the point has been recorded for the first time. (c) Corresponding front-view camera image, where we highlight the traffic signs. (d) Corresponding relative translation error plot for each time step. The dots are the calculated relative translational errors in each time stamp and the curves are polynomial fitting results of those dots.}
 \label{fig:highway-result}
 \vspace{-0.4cm}
\end{figure}

Our experimental evaluation is designed to support our main claims that we are (i) able to accurately map even in situations with a considerable amount of moving objects
and we are (ii) able to achieve better performance than simply removing possibly moving objects in general environments, including urban, countryside, and highway scenes.

To this end, we evaluate our approach using data from the KITTI Vision Benchmark \cite{geiger2012cvpr}, where we use the provided point clouds generated by a Velodyne HDL-64E S2 recorded at a rate of $10$\,Hz.
To evaluate the performance of odometry, the dataset proposes to compute relative errors in respect to translation and rotation averaged over different distances between poses and averaging it.
The ground truth poses are generated using pose information from an inertial navigation system and in most sequences the GPS position is referenced to a base station, which makes it quite accurate, but still often only locally consistent.

In the following, we compare our proposed approach (denoted by SuMa++) against the original surfel-based mapping (denoted by SuMa), and SuMa with the naive approach of removing all movable classes (cars, buses, trucks, bicycles, motorcycles, other vehicles, persons, bicyclists, motorcyclist) given by the semantic segmentation (denoted by SuMa\_nomovable).

The RangeNet++ for the semantic segmentation was trained using point-wise annotations~\cite{behley2019iccv} using all training sequences from the KITTI Odometry Benchmark, which are the labels available for training purposes. This includes sequences $00$ to $10$, except for sequence $08$ which is left out for validation.

We tested our approach on an Intel Xeon(R) W-2123 with 8 cores @3.60 GHz with 16 GB RAM, and an Nvidia Quadro P4000 with 8 GB RAM.
The RangeNet++ needs on average $75\,$ms to generate point-wise labels for each scan and the surfel-mapping needs on average $48\,$ms, but we need at most $190\,$ms to integrate loop closures in some situations (on sequence $00$ of the training set with multiple loop closures).

\begin{table}[t]
\centering
\vspace{0.2cm}
\caption{Results on KITTI Road dataset}
\scalebox{0.9}{
\scriptsize{
\begin{tabular}{ccccc}
\toprule
Sequence & Environment & \multicolumn{3}{c}{Approach} \\
\cline{3-5}
 &  & SuMa & SuMa\_nomovable & SuMa++ \Tstrut \\
\midrule
30 & country & $0.38$/$0.96$ & $0.39$/$0.97$ & $\mathbf{0.38}$/$\mathbf{0.90}$ \\
31 & country & $1.54$/$2.02$ & $1.66$/$2.13$ & $\mathbf{1.19}$/$\mathbf{2.02}$ \\
32 & country & $1.38$/$1.70$ & $1.63$/$1.76$ & $\mathbf{1.00}$/$\mathbf{1.57}$ \\
33 & highway & $\mathbf{1.61}$/$\mathbf{1.79}$ & $1.72$/$1.80$ & $1.67$/$1.87$ \\
34 & highway & $0.79$/$1.17$ & $0.70$/$1.14$ & $\mathbf{0.60}$/$\mathbf{1.09}$ \\
35 & highway & $5.11$/$26.8$ & $3.20$/$1.22$ & $\mathbf{2.90}$/$\mathbf{1.11}$ \\
36 & highway & $0.93$/$1.31$ & $0.95$/$\mathbf{1.30}$ & $\mathbf{0.93}$/$1.40$ \\
37 & country & $0.65$/$1.51$ & $0.62$/$\mathbf{1.36}$ & $\mathbf{0.60}$/$1.48$ \\
38 & highway & $1.07$/$1.66$ & $1.04$/$1.46$ & $\mathbf{0.89}$/$\mathbf{1.42}$ \\
39 & country & $0.46$/$1.04$ & $0.47$/$\mathbf{0.98}$ & $\mathbf{0.44}$/$1.05$ \\
40 & country & $1.09$/$18.0$ & $0.79$/$\mathbf{1.92}$ & $\mathbf{0.75}$/$1.95$ \\
41 & highway & $1.24$/$15.6$ & $\mathbf{0.92}$/$\mathbf{1.46}$ & $1.14$/$1.67$ \\
\midrule
\multicolumn{2}{c}{Average} & $1.35$/$6.13$ & $1.17$/$1.46$ & $\mathbf{1.04}$/$\mathbf{1.46}$\\
\midrule

\bottomrule
\multicolumn{5}{p{\linewidth}}{\vspace{0.01cm}Relative errors averaged over trajectories of $5$ to $400$\,m length: relative rotational error in $\mathrm{degrees}$ per $100$\,m / relative translational error in $\%$. Bold numbers indicate top performance for laser-based approaches.\newline  We rename the KITTI road raw dataset '2011\_09\_26\_drive\_0015\_sync'-'2011\_10\_03\_drive\_0047\_sync' into sequences 30-41.} \\
\end{tabular}
}
}
\label{tab:roads-set}
\vspace{-0.4cm}
\end{table}

\subsection{KITTI Road Sequences}

\begin{figure*}[t]
\vspace{0.2cm}
  \centering
 \subfigure[Sequence 32]{\includegraphics[width=0.23\linewidth]{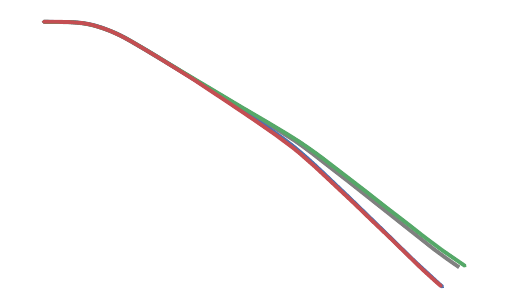}}
 \subfigure[Sequence 35]{\includegraphics[width=0.23\linewidth]{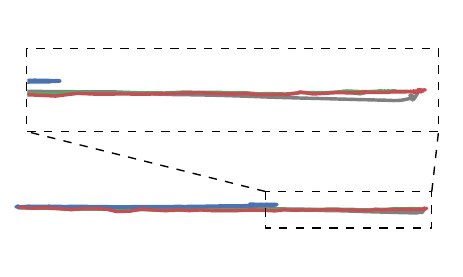}}
 \subfigure[Sequence 40]{\includegraphics[width=0.23\linewidth]{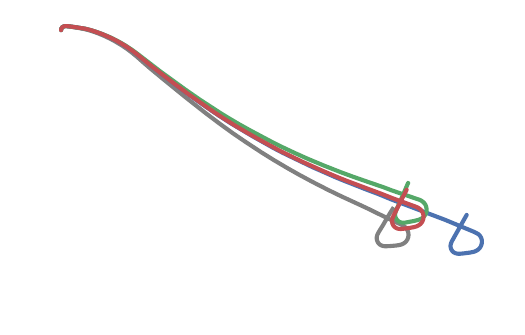}}
 \subfigure[Sequence 41]{\includegraphics[width=0.23\linewidth]{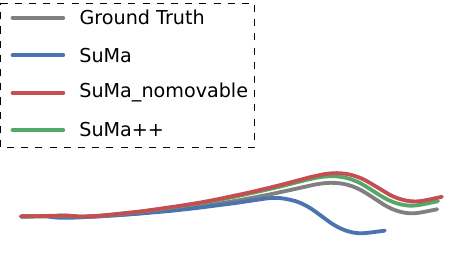}}
 \caption{Trajectories of different methods test on KITTI road dataset.}
\label{fig:kitti_road_trajs}
\vspace{-0.4cm}
\end{figure*}

\begin{table*}[t]
\centering
\vspace{0.2cm}
\caption{Results on KITTI Odometry (training)}
\scalebox{0.85}{
\scriptsize{
\begin{tabular}{lcccccccccccc}
\toprule
& \multicolumn{11}{c}{Sequence} \\
Approach & 00* & 01 & 02* & 03 & 04 & 05* & 06* & 07* & 08* & 09* & 10 & Average \\
  & urban & highway & urban & country & country & country & urban & urban & urban & urban & country &  \\
\midrule

SuMa & $0.23$/$0.68$ & $0.54$/$1.70$ & $0.48$/$1.20$ & $0.50$/$0.74$ & $0.27$/$0.44$ & $0.20$/$0.43$ & $0.3$/$0.54$ & $0.54$/$0.74$ & $0.38$/$1.20$ & $0.22$/$0.62$ & $0.32$/$0.72$  & $0.36$/$0.83$ \\
\midrule

SuMa\_nomovable & $22.0$/$58.0$ & $0.57$/$1.70$ & $25.0$/$63.0$ & $0.45$/$0.67$ & $0.26$/$0.37$ & $14.0$/$36.0$ & $0.22$/$0.47$ & $0.21$/$0.34$ & $13.0$/$32.0$ & $13.0$/$45.0$ & $12.0$/$19.0$  & $23.3$/$9.24$ \\
\midrule

SuMa++ & $0.22$/$0.64$ & $0.46$/$1.60$ & $0.37$/$1.00$ & $0.46$/$\mathbf{0.67}$ & $0.26$/$0.37$ & $0.20$/$0.40$ & $0.21$/$0.46$ & $0.19$/$0.34$ & $0.35$/$1.10$ & $0.23$/$\mathbf{0.47}$ & $0.28$/$0.66$  & $0.29$/$0.70$ \\
\midrule
\midrule

 IMLS-SLAM \cite{deschaud2018icra} & \hspace{1em}-/$\mathbf{0.50}$ & \hspace{1em}-/$\mathbf{0.82}$ & \hspace{1em}-/$\mathbf{0.53}$ & \hspace{1em}-/$0.68$ & \hspace{1em}-/$\mathbf{0.33}$ & \hspace{1em}-/$\mathbf{0.32}$ & \hspace{1em}-/$\mathbf{0.33}$ & \hspace{1em}-/$\mathbf{0.33}$ & \hspace{1em}-/$\mathbf{0.80}$ & \hspace{1em}-/$0.55$ & \hspace{1em}-/$\mathbf{0.53}$ & \hspace{1em}-/$\mathbf{0.55}$ \\
 \midrule
 
 LOAM \cite{zhang2017auro} & \hspace{1em}-/$0.78$ & \hspace{1em}-/$1.43$ & \hspace{1em}-/$0.92$ & \hspace{1em}-/$0.86$ & \hspace{1em}-/$0.71$ & \hspace{1em}-/$0.57$ & \hspace{1em}-/$0.65$ & \hspace{1em}-/$0.63$ & \hspace{1em}-/$1.12$ & \hspace{1em}-/$0.77$ & \hspace{1em}-/$0.79$ & \hspace{1em}-/$0.84$ \\
\bottomrule
\multicolumn{12}{p{\linewidth}}{\vspace{0.01cm}Relative errors averaged over trajectories of $100$ to $800$\,m length: relative rotational error in $\mathrm{degrees}$ per $100$\,m / relative translational error in $\%$. \newline Sequences marked with an asterisk contain loop closures. Bold numbers indicate best performance in terms of translational error.} \\
\end{tabular}
}
}
\label{tab:training-set}
\vspace{-0.4cm}
\end{table*}

The first experiment is designed to show that our approach is able to generate consistent maps even in situations with many moving objects. We show results on sequences from the road category of the raw data of the KITTI Vision Benchmark. Note that these sequences are not part of the odometry benchmark, and therefore no labels are provided for the semantic segmentation, meaning that our network learned to infer the semantic classes of road driving scenes, and it is not simply memorizing.
These sequences, especially the highway sequences, are challenging for SLAM methods, since here most of the objects are moving cars. Moreover, there are only sparse distinct features on the side of the road, like traffic signs or poles.
Building corners or other more distinctive features are not available to guide the registration process.
In such situations, wrong correspondences on consistently moving outliers (like cars in a traffic jam) often lead to wrongly estimated pose changes and consequently inconsistencies in the generated map.

\figref{fig:highway-result} shows an example generated with SuMa and the proposed SuMa++.
In the case of the purely geometric approach, we clearly see that the pose cannot be correctly estimated, since the highlighted traffic signs show up at different locations leading to large inconsistencies.
With our proposed approach, where we are able to correctly filter the moving cars, we instead generate a consistent map as evident by the highlighted consistently mapped traffic signs. We also plot the relative translational errors of odometry results of both SuMa and SuMa++ in this example. The dots represent the  relative translational errors in each timestamp and the curves are polynomial fitting results given the dots. It shows that SuMa++ achieves more accurate pose estimates in such a challenging environments with many outliers caused by moving objects.

\tabref{tab:roads-set} shows the relative translational and relative rotational error and \figref{fig:kitti_road_trajs} shows the corresponding trajectories for different methods tested on this part of the dataset.
Generally, we see that our proposed approach, SuMa++, generates more consistent trajectories and achieves in most cases a lower translational error than SuMa.
Compared to the baseline of just removing all possibly moving objects, SuMa\_nomovable, we see very similar performance compared to SuMa++.
This confirms that the major reason for the worse performance of SuMa in such cases is the inconsistencies caused by actually moving objects.
However, we will show in the next experiments that removing all potentially moving objects can also have negative effects on the pose estimation performance in urban environments.

\subsection{KITTI Odometry Benchmark}

The second experiment is designed to show that our methods performs superior compared to simply removing certain semantic classes from the observations. This evaluation is performed on the KITTI Odometry Benchmark.

\tabref{tab:training-set} shows the relative translational and relative rotational errors. IMLS-SLAM~\cite{deschaud2018icra} and LOAM~\cite{zhang2017auro} are state-of-the-art LiDAR-based SLAM approaches.
In most sequences, we can see similar performance of SuMa++ compared to the state-of-the-art.
More interestingly, the baseline method SuMa\_nomovable diverges, particularly in urban scenes. 

This might be counter-intuitive since these environments contain a considerable amount of man-made structures and other more distinctive features.
But there are two reasons contributing to this worse performance that become clear when one looks at the results and the configuration of the scenes where mapping errors occur. 
First, even though we try to improve the results of the semantic segmentation, there are wrong predictions that lead to a removal of surfels in the map that are actually static.
Second, the removal of parked cars is problems as these are good and distinctive features for aligning scans.
Both effects contribute to making the surfel map sparser. This is even more critical as parked cars are the only distinctive or reliable features.
In conclusion, the simple removal of certain classes is at least in our situation sub-optimal and can lead to worse performance.

To evaluate the performance of our approach in unseen trajectories, we uploaded our results for server-side evaluation on unknown KITTI test sequences so that no parameter tuning on the test set is possible. Thus, this serves as a good proxy for the real-world performance of our approach. In the test set, we achieved an average rotational error of $0.0032$ deg/m and an average translational error of $1.06$\%, which is an improvement in terms of translational error, when compared to $0.0032$ deg/m and $1.39$\% of the original SuMa.

\subsection{Discussion}

During the map updating process, we only penalize surfels of dynamics of movable objects, which means we do not penalize semantically static objects, e.g. vegetation, even though sometimes leaves of vegetation change and the appearance of vegetation changes with the viewpoint due to laser beams that only get reflected from certain viewpoints.
Our motivation for this is that they can also serve as good landmarks, e.g., the trunk of a tree is static and a good feature for pose estimation.
Furthermore, the original geometric-based outlier rejection mechanism employing the Huber norm often down-weights such parts. 
	
There is an obvious limitation of our method: we cannot filter out dynamic objects in the first observation. 
Once there is a large number of moving objects in the first scan, our method will fail because we cannot estimate a proper initial velocity or pose. We solve this problem by removing all potentially movable object classes in the initialization period. 
However, a more robust method would be to backtrack changes due to the change in the observed moving state and thus update the map retrospectively.
	
Lastly, the results of our second experiment shows convincingly that blindly removing a certain set of classes can deteriorate the localization accuracy, but potentially moving objects still might be removed from a long-term representation of a map to also allow for a  representation of otherwise occluded parts of the environment that might be visible at a different point of time.
 
\section{Conclusion}
\label{sec:conclusion}

In this paper, we presented a novel approach to build semantic maps enabled by a laser-based semantic segmentation of the point cloud not requiring any camera data. We exploit this information to improve pose estimation accuracy in otherwise ambiguous and challenging situations.
In particular, our method exploits semantic consistencies between scans and the map to filter out dynamic objects and provide higher-level constraints during the ICP process.
This allows us to successfully combine semantic and geometric information based solely on three-dimensional laser range scans to achieve considerably better pose estimation accuracy than the pure geometric approach.
We evaluated our approach on the KITTI Vision Benchmark dataset showing the advantages of our approach in comparison to purely geometric approaches. 
Despite these encouraging results, there are several avenues for future research on semantic mapping.
In future work, we plan to investigate the usage of semantics for loop closure detection and the estimation of more fine-grained semantic information, such as lane structure or road type.


\bibliographystyle{plain}

\bibliography{glorified,new}

\begin{thebibliography}{10}

\bibitem{behley2019iccv}
J.~Behley, M.~Garbade, A.~Milioto, J.~Quenzel, S.~Behnke, C.~Stachniss, and
  J.~Gall.
\newblock {SemanticKITTI: A Dataset for Semantic Scene Understanding of LiDAR
  Sequences}.
\newblock In {\em Proc. of the IEEE/CVF International Conf.~on Computer Vision
  (ICCV)}, 2019.

\bibitem{behley2018rss}
J.~Behley and C.~Stachniss.
\newblock {Efficient Surfel-Based SLAM using 3D Laser Range Data in Urban
  Environments}.
\newblock In {\em Proc.~of Robotics: Science and Systems (RSS)}, 2018.

\bibitem{bescos2018ral}
B.~Bescos, J.M. Fácil, J.~Civera, and J.~Neira.
\newblock {DynaSLAM: Tracking, Mapping, and Inpainting in Dynamic Scenes}.
\newblock {\em IEEE Robotics and Automation Letters (RA-L)}, 3(4):4076--4083,
  2018.

\bibitem{bowman2017icra}
S.~Bowman, N.~Atanasov, K.~Daniilidis, and G.J. Pappas.
\newblock {Probabilistic Data Association for Semantic SLAM}.
\newblock In {\em Proc.~of the IEEE Intl.~Conf.~on Robotics \& Automation
  (ICRA)}, 2017.

\bibitem{brasch2018iros}
N.~Brasch, A.~Bozic, J.~Lallemand, and F.~Tombari.
\newblock {Semantic Monocular SLAM for Highly Dynamic Environments}.
\newblock In {\em Proc.~of the IEEE/RSJ Intl.~Conf.~on Intelligent Robots and
  Systems (IROS)}, pages 393--400, 2018.

\bibitem{cadena2016tro}
C.~Cadena, L.~Carlone, H.~Carrillo, Y.~Latif, D.~Scaramuzza, J.~Neira, I.~Reid,
  and J.J. Leonard.
\newblock {Past, Present, and Future of Simultaneous Localization And Mapping:
  Towards the Robust-Perception Age}.
\newblock {\em IEEE Trans.~on Robotics (TRO)}, 32:1309--1332, 2016.

\bibitem{deschaud2018icra}
J.~Deschaud.
\newblock Imls-slam: scan-to-model matching based on 3d data.
\newblock In {\em Proc.~of the IEEE Intl.~Conf.~on Robotics \& Automation
  (ICRA)}, 2018.

\bibitem{dube2018rss}
R.~Dub\'e, A.~Cramariuc, D.~Dugas, J.~Nieto, R.~Siegwart, and C.~Cadena.
\newblock {SegMap: 3D Segment Mapping using Data-Driven Descriptors}.
\newblock In {\em Proc.~of Robotics: Science and Systems (RSS)}, 2018.

\bibitem{ganti2018arxiv}
P.~Ganti and S.L. Waslander.
\newblock {Network Uncertainty Informed Semantic Feature Selection for Visual
  SLAM}.
\newblock {\em arXiv preprint}, 2018.

\bibitem{geiger2012cvpr}
A.~Geiger, P.~Lenz, and R.~Urtasun.
\newblock {Are we ready for Autonomous Driving? The KITTI Vision Benchmark
  Suite}.
\newblock In {\em Proc.~of the IEEE Conf.~on Computer Vision and Pattern
  Recognition (CVPR)}, pages 3354--3361, 2012.

\bibitem{jeong2018esa}
J.~Jeong, T.~S. Yoon, and J.~B. Park.
\newblock {Multimodal Sensor-Based Semantic 3D Mapping for a Large-Scale
  Environment}.
\newblock {\em Expert Systems with Applications}, 105:1--10, 2018.

\bibitem{jeong2018sensors}
J.~Jeong, T.S. Yoon, and J.B. Park.
\newblock {Towards a Meaningful 3D Map Using a 3D Lidar and a Camera}.
\newblock {\em IEEE Sensors Journal}, 18(8), 2018.

\bibitem{kuemmerle2013icra}
R.~K\"ummerle, M.~Ruhnke, B.~Steder, C.~Stachniss, and W.~Burgard.
\newblock {A Navigation System for Robots Operating in Crowded Urban
  Environments}.
\newblock In {\em Proc.~of the IEEE Intl.~Conf.~on Robotics \& Automation
  (ICRA)}, Karlsruhe, Germany, 2013.

\bibitem{li2018eccv}
P.~Li, T.~Qin, and S.~Shen.
\newblock {Stereo Vision-based Semantic 3D Object and Ego-motion Tracking for
  Autonomous Driving}.
\newblock In {\em Proc.~of the Europ.~Conf.~on Computer Vision (ECCV)}, 2018.

\bibitem{li2017cvpr}
X.~Li, Z.~Liu, P.~Luo, C.C. Loy, and X.~Tang.
\newblock {Not All Pixels Are Equal: Difficulty-Aware Semantic Segmentation via
  Deep Layer Cascade}.
\newblock In {\em Proc.~of the IEEE Conf.~on Computer Vision and Pattern
  Recognition (CVPR)}, 2017.

\bibitem{liang2018eccv}
M.~Liang, B.~Yang, S.~Wang, and R.~Urtasun.
\newblock Deep continuous fusion for multi-sensor 3d object detection.
\newblock In {\em Proc.~of the Europ.~Conf.~on Computer Vision (ECCV)}, pages
  641--656, 2018.

\bibitem{lianos2018eccv}
K.N. Lianos, J.L. Schonberger, M.~Pollefeys, and T.~Sattler.
\newblock {VSO: Visual Semantic Odometry}.
\newblock In {\em Proc.~of the Europ.~Conf.~on Computer Vision (ECCV)}, pages
  234--250, 2018.

\bibitem{mccormac2018threedv}
J.~McCormac, R.~Clark, M.~Bloesch, A.~Davison, and S.~Leutenegger.
\newblock {Fusion++: Volumetric Object-Level SLAM}.
\newblock In {\em Proc.~of the Intl.~Conf.~on 3D Vision (3DV)}, pages 32--41,
  2018.

\bibitem{mccormac2017icra}
J.~McCormac, A.~Handa, A.J. Davison, and S.~Leutenegger.
\newblock {SemanticFusion: Dense 3D Semantic Mapping with Convolutional Neural
  Networks}.
\newblock In {\em Proc.~of the IEEE Intl.~Conf.~on Robotics \& Automation
  (ICRA)}, 2017.

\bibitem{milioto2019iros}
A.~Milioto, I.~Vizzo, J.~Behley, and C.~Stachniss.
\newblock {RangeNet++: Fast and Accurate LiDAR Semantic Segmentation}.
\newblock In {\em Proceedings of the IEEE/RSJ Int. Conf. on Intelligent Robots
  and Systems (IROS)}, 2019.

\bibitem{palazzolo2019iros}
E.~Palazzolo, J.~Behley, P.~Lottes, P.~Giguere, and C.~Stachniss.
\newblock { ReFusion: 3D Reconstruction in Dynamic Environments for RGB-D
  Cameras Exploiting Residuals}.
\newblock In {\em Proc.~of the IEEE/RSJ Intl.~Conf.~on Intelligent Robots and
  Systems (IROS)}, 2019.

\bibitem{parkison2018bmvc}
S.A. Parkison, L.~Gan, M.G. Jadidi, and R.M. Eustice.
\newblock {Semantic Iterative Closest Point through Expectation-Maximization}.
\newblock In {\em Proc.~of British Machine Vision Conference (BMVC)}, page 280,
  2018.

\bibitem{pomerleau2014icra}
F.~Pomerleau, P.~Kr{\"u}siand, F.~Colas, P.~Furgale, and R.~Siegwart.
\newblock Long-term 3d map maintenance in dynamic environments.
\newblock In {\em Proc.~of the IEEE Intl.~Conf.~on Robotics \& Automation
  (ICRA)}, 2014.

\bibitem{redmon2018arxiv}
J.~Redmon and A.~Farhadi.
\newblock {YOLOv3: An Incremental Improvement}.
\newblock {\em arXiv preprint}, 2018.

\bibitem{ruenz2018ismar}
M.~R{\"u}nz, M.~Buffier, and L.~Agapito.
\newblock {MaskFusion: Real-Time Recognition, Tracking and Reconstruction of
  Multiple Moving Objects}.
\newblock In {\em Proc.~of the Intl.~Symposium~on Mixed and Augmented Reality
  (ISMAR)}, pages 10--20, 2018.

\bibitem{salas2013cvpr}
R.~F. Salas-Moreno, R.~A. Newcombe, H.~Strasdat, P.~HJ Kelly, and A.~J.
  Davison.
\newblock Slam++: Simultaneous localisation and mapping at the level of
  objects.
\newblock In {\em Proc.~of the IEEE Conf.~on Computer Vision and Pattern
  Recognition (CVPR)}, pages 1352--1359, 2013.

\bibitem{stachniss2016handbook-slamchapter}
C.~Stachniss, J.~Leonard, and S.~Thrun.
\newblock {\em {Springer Handbook of Robotics, 2nd edition}}, chapter
  Chapt.~46: Simultaneous Localization and Mapping.
\newblock Springer Verlag, 2016.

\bibitem{suenderhauf2017iros}
N.~Suenderhauf, T.~Pham, Y.~Latif, M.J. Milford, and I.~Reid.
\newblock {Meaningful Maps with Object-Oriented Semantic Mapping}.
\newblock In {\em Proc.~of the IEEE/RSJ Intl.~Conf.~on Intelligent Robots and
  Systems (IROS)}, 2017.

\bibitem{sun2018ral}
L.~Sun, Z.~Yan, A.~Zaganidis, C.~Zhao, and T.~Duckett.
\newblock {Recurrent-OctoMap: Learning State-Based Map Refinement for Long-Term
  Semantic Mapping With 3-D-Lidar Data}.
\newblock {\em IEEE Robotics and Automation Letters (RA-L)}, 3(4):3749--3756,
  2018.

\bibitem{tateno2017cvpr}
K.~Tateno, F.~Tombari, I.~Laina, and N.~Navab.
\newblock {CNN-SLAM: Real-time dense monocular SLAM with learned depth
  prediction}.
\newblock In {\em Proc.~of the IEEE Conf.~on Computer Vision and Pattern
  Recognition (CVPR)}, volume~2, pages 6565--6574, 2017.

\bibitem{thrun2005probrobbook}
S.~Thrun, W.~Burgard, and D.~Fox.
\newblock {\em {Probabilistic Robotics}}.
\newblock MIT Press, 2005.

\bibitem{vineet2015icra}
V.~Vineet, O.~Miksik, M.~Lidegaard, M.~Niessner, S.~Golodetz, V.~Prisacariu,
  O.~Kahler, D.~Murray, S.~Izadi, P.~Perez, and P.~Torr.
\newblock {Incremental Dense Semantic Stereo Fusion for Large-Scale Semantic
  Scene Reconstruction}.
\newblock In {\em Proc.~of the IEEE Intl.~Conf.~on Robotics \& Automation
  (ICRA)}, 2015.

\bibitem{vysotska2016ral}
O.~Vysotska and C.~Stachniss.
\newblock {Lazy Data Association for Image Sequences Matching under Substantial
  Appearance Changes}.
\newblock {\em IEEE Robotics and Automation Letters (RA-L)}, 2016.

\bibitem{wang2017iros}
J.~Wang and J.~Kim.
\newblock {Semantic segmentation of urban scenes with a location prior map
  using lidar measurements}.
\newblock In {\em Proc.~of the IEEE/RSJ Intl.~Conf.~on Intelligent Robots and
  Systems (IROS)}, 2017.

\bibitem{wu2018icra-scnn}
B.~Wu, A.~Wan, X.~Yue, and K.~Keutzer.
\newblock {SqueezeSeg: Convolutional Neural Nets with Recurrent CRF for
  Real-Time Road-Object Segmentation from 3D LiDAR Point Cloud}.
\newblock In {\em Proc.~of the IEEE Intl.~Conf.~on Robotics \& Automation
  (ICRA)}, 2018.

\bibitem{yan2014threedv}
J.~Yan, D.~Chen, H.~Myeong, T.~Shiratori, and Y.~Ma.
\newblock {Automatic Extraction of Moving Objects from Image and LIDAR
  Sequences}.
\newblock In {\em Proc.~of the Intl.~Conf.~on 3D Vision (3DV)}, 2014.

\bibitem{yang2017iros}
S.~Yang, Y.~Huang, and S.~Scherer.
\newblock {Semantic 3D occupancy mapping through efficient high order CRFs}.
\newblock In {\em Proc.~of the IEEE/RSJ Intl.~Conf.~on Intelligent Robots and
  Systems (IROS)}, 2017.

\bibitem{yu2018iros}
C.~Yu, Z.~Liu, X.~Liu, F.~Xie, Y.~Yang, Q.~Wei, and Q.~Fei.
\newblock {DS-SLAM: A Semantic Visual SLAM towards Dynamic Environments}.
\newblock In {\em Proc.~of the IEEE/RSJ Intl.~Conf.~on Intelligent Robots and
  Systems (IROS)}, pages 1168--1174, 2018.

\bibitem{zaganidis2018ral}
A.~Zaganidis, L.~Sun, T.~Duckett, and G.~Cielniak.
\newblock {Integrating Deep Semantic Segmentation Into 3-D Point Cloud
  Registration}.
\newblock {\em IEEE Robotics and Automation Letters (RA-L)}, 3(4):2942--2949,
  2018.

\bibitem{zhang2017auro}
J.~Zhang and S.~Singh.
\newblock {Low-drift and real-time lidar odometry and mapping}.
\newblock {\em Autonomous Robots}, 41:401--416, 2017.

\end{thebibliography}

\end{document}